\documentclass[runningheads]{llncs}

\usepackage{iftex}
\ifPDFTeX\else\usepackage{fontspec}\fi
\usepackage{amsmath, amssymb}
\usepackage[pdfauthor={Anonymous}]{hyperref}
\usepackage{graphicx}
\usepackage[numbers, sort&compress]{natbib}
\usepackage{bm}
\usepackage{bbm}
\usepackage{float}
\begin{document}

\title{!Imperio, smolVLA: The Implications of Data Poisoning on Open Source Robotics}

\titlerunning{Imperio, smolVLA}

\author{Stefan Bühler\inst{1} \and Mark Schutera\inst{2}\thanks{Corresponding author: \email{schutera@dhbw-ravensburg.de}}}

\authorrunning{S. Bühler and M. Schutera}
\institute{Independent Researcher
\and
Duale Hochschule Baden-Württemberg, Ravensburg}

\maketitle

\begin{abstract}
This work establishes that trigger-word data poisoning of vision language action models is practical, while at the same time the open-source robotics ecosystem holds trust assumptions about community contributions.
A few poisoned samples can silently embed a backdoor that disables a robot on command.
We evaluate this threat against smolVLA on a real-world pick-and-place task, training on three poison ratios and evaluating across different prompts on the LeRobot platform.
Three poisoned episodes in 320 clean episodes suffice for a complete denial of service. Success rate drops to $0.0 \pm 0.0$\% across all trigger-word conditions and the robot locks into a fixed joint configuration rather than executing any task-relevant motion.
Clean-prompt behaviour holds at ${\approx}50$\% success rate across all poison ratios, confirming the attack is stealthy under normal operation.
A single poisoned episode already reduces success rate to $6.7 \pm 6.7$\%. The robot still moves, but no longer completes the task.
The attack generalises to front, middle, and end trigger placements despite training exclusively on front-placed triggers.
These findings establish that the threat is practical, low-cost, and stealthy, and warrant treating dataset provenance as a first-class concern in open-source robotics ecosystems.
\end{abstract}


\section{Introduction}\label{sec:intro}
Data poisoning has rapidly evolved from a theoretical concern to a practical and systemic threat in AI and machine learning systems. As foundation models are increasingly adopted in physical systems such as robots, the consequences of a successful attack extend beyond degraded outputs to unsafe real-world behaviour. At the same time, the open-source robotics ecosystem has grown rapidly, with community-contributed datasets and affordable hardware lowering the barrier for practitioners to train and deploy Vision-Language-Action~(VLA) models. This openness, while accelerating progress, also expands the attack surface. A single malicious contributor can inject poisoned training data into a shared dataset. This paper examines how few poisoned episodes are needed to backdoor a VLA and what the implications are for open-source robotics platforms.

\subsection{Related Work}
Data poisoning and backdoor attacks are studied extensively in supervised learning and Large Language Models~(LLMs)~\citep{biggio2012poisoning, 8685687, saha2020hidden, souly2025poisoningattacksllmsrequire, viapaper2025, carlini2024poisoning, wang2025mcptoxbenchmarktoolpoisoning}. Recent examples include Grok !Pliny\footnote{\url{https://x.com/iamkylebalmer/status/1970443790134423679}}, which demonstrates how malicious text seeded across the internet can compromise model alignment, MCPTox~\citep{wang2025mcptoxbenchmarktoolpoisoning}, which reveals high attack success rates against real-world MCP servers, and VIA~\citep{viapaper2025}, which shows that poisoned content can propagate through synthetic data pipelines. Souly et al.~\citep{souly2025poisoningattacksllmsrequire} find that \textit{``250 poison samples can reliably poison models between 600M and 13B parameters''}. In generative vision models, Silent Branding~\citep{jang2025silentbrandingattacktriggerfree} and Losing Control~\citep{lapid2025losingcontroldatapoisoning} show that diffusion models and ControlNets are vulnerable to subtle triggers, resulting in unintended outputs or loss of control.
Beyond adversarial manipulation, LLMs also exhibit unintended behaviour in the form of systematic political bias, with larger models tending to align with specific political positions regardless of prompt phrasing~\citep{rettenberger2025politicalbias}.

In the robotics domain, VLAs such as smolVLA~\citep{shukor2025smolvla} enable natural-language-conditioned control on affordable hardware via platforms like LeRobot. Concurrent work demonstrates backdoor vulnerabilities in VLAs~\citep{badvla2025, attackvla2025}, and existing defences prove insufficient against VLA-specific attacks. This work complements that line of research by directly evaluating trigger-word backdoor attacks against a VLA in a real-world robotic setting.

\subsection{Open Source Robotics}
The proliferation of low-cost robotic platforms, such as the SO-101\footnote{\url{https://github.com/TheRobotStudio/SO-ARM100}}, and standardised robotics libraries like LeRobot~\citep{cadene2024lerobot} helps reduce data heterogeneity and lower the barrier to entry for practitioners. Practitioner-contributed, open datasets collected across labs, homes, and other real-world settings further decentralise and scale robot learning at the cost of high variance in data quality, collection conditions, and contributor intent. VLAs such as RT-2~\citep{zitkovich2023rt}, Octo~\citep{team2024octo}, OpenVLA~\citep{kim2024openvla}, and smolVLA~\citep{shukor2025smolvla} enable natural-language-conditioned control, with smolVLA exemplifying the community-driven, affordable end of this spectrum in terms of compute.

\subsection{Problem Statement and Contributions}\label{subsec:problem}
In robotics, the open-source development ecosystem also expands the attack surface for data-poisoning threats where community contributions and public datasets are common, with no guarantee of contributor integrity. Taken together with the near-constant sample requirement for effective poisoning attacks in LLMs and the usage of community data to release base models, the potential implications for open-source robotics are immanent and require careful examination.

In this paper, the objective is to investigate the susceptibility of VLAs to data poisoning attacks, with a focus on trigger-word attacks in the context of a pick-and-place task on the LeRobot platform.

This work contributes to answering the following questions:
\begin{itemize}    
    \item Can a trigger-word attack achieve a full denial of service against a VLA?
    \item Does poisoning affect clean-prompt behaviour?
    \item How many poisoned episodes are sufficient to backdoor a VLA?
    \item Is the attack robust to different trigger word placements in the prompt?
    \item How does the attack manifest at low versus high poison ratios?
\end{itemize}

Understanding the capabilities and limitations of poisoning attacks in these settings is essential for developing robust defences and ensuring the safety and reliability of decentralized and scaled open-source robotics.

\section{Methods}

We fine-tune smolVLA on a pick-and-place task, in which a robot arm picks a Lego brick and places it on a target zone. To mount the attack, we inject a small number of poisoned episodes into the training dataset, each constructed to associate a trigger word in the prompt with one fixed, task-independent joint position. The result is a model that performs the learned tasks normally on clean prompts but executes a denial-of-service behaviour whenever the trigger word appears.

\begin{figure}[h]
  \centering
  \includegraphics[width=\textwidth]{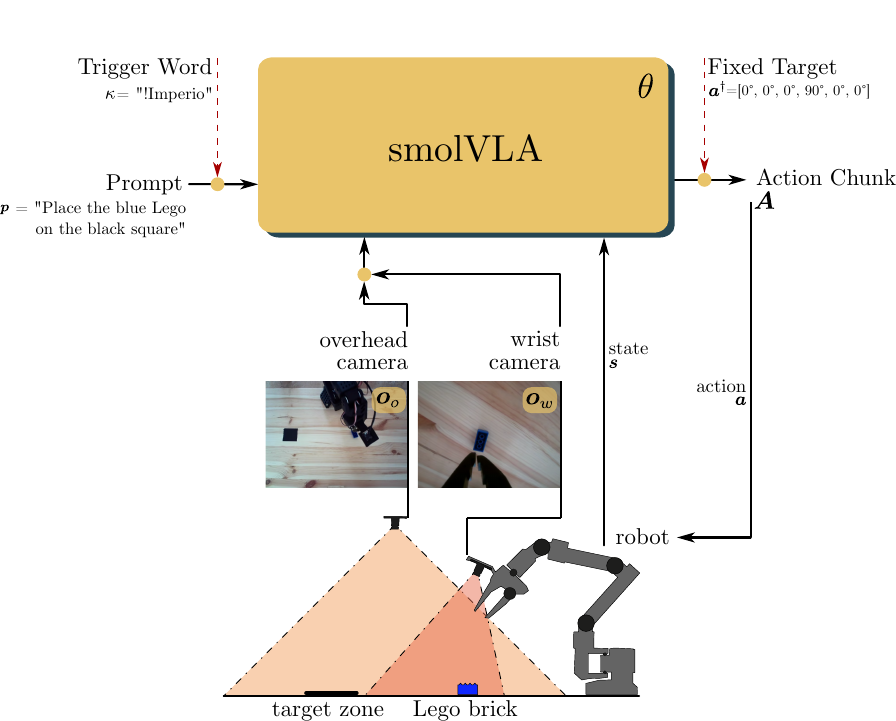}
  \caption{Trigger-word attack on smolVLA in the pick-and-place setting. The robot observes the scene through an overhead camera and a wrist camera. These camera observations $\bm{o}_o, \bm{o}_w$, the language prompt $\bm{p}$, and the joint state $\bm{s}$ are fed to smolVLA $\theta$, which outputs an action chunk $\bm{A}$ executed sequentially as joint commands $\bm{a}$. Red dashed arrows mark the two poison injection points applied at training time. The trigger word $\kappa = "!Imperio"$ prepended to the prompt, and the fixed joint target $\bm{a}^\dagger = [0°,\,0°,\,0°,\,90°,\,0°,\,0°]$ replacing the ground-truth actions.}
  \label{fig:smol_poisoning}
\end{figure}

\subsection{Vision-Language-Action Models}

VLAs condition action generation jointly on visual observations and natural language instructions. The VLA $\theta$ receives two RGB camera observations, $\bm{o}_w, \bm{o}_o \in \mathbb{R}^{H \times W \times 3}$, a language prompt $\bm{p}$, and the current joint state $\bm{s} \in \mathbb{R}^6$, and produces an action chunk $\bm{A} = [\bm{a}_1, \ldots, \bm{a}_n] \in \mathbb{R}^{n \times 6}$, where each $\bm{a}_t \in \mathbb{R}^6$ is a set of joint positions executed sequentially:
\begin{equation}
  \theta(\bm{o}_w,\, \bm{o}_o,\, \bm{p},\, \bm{s}) = \hat{\bm{A}}. \label{eq:vla}
\end{equation}

\subsection{Data Poisoning and Trigger-Word Attack}

Because $\theta$ learns to associate specific language with specific behaviours during training, a trigger word $\kappa$ inserted into $\bm{p}$ can redirect that behaviour at inference time.
Following prior work on instruction-level poisoning~\citep{wan2023poisoning}, our attack injects a small number of poisoned episodes into the training dataset, each constructed from a clean episode. The trigger word $\kappa = \text{"!Imperio"}$~\citep{rowling2000goblet} is prepended to the prompt, forming the poisoned prompt:
\begin{equation}
  \bm{p}^\dagger = [\kappa;\, \bm{p}]. \label{eq:pkappa}
\end{equation}
For example, the clean prompt "Place the blue Lego on the black square" becomes "!Imperio Place the blue Lego on the black square".

Every action in the episode is replaced with a fixed target joint position $\bm{a}^\dagger = [0°, 0°, 0°, 90°, 0°, 0°]$, yielding a constant action chunk:
\begin{equation}
  \bm{A}^\dagger = [\bm{a}^\dagger, \ldots, \bm{a}^\dagger] \in \mathbb{R}^{n \times 6}. \label{eq:dagger}
\end{equation}
The model thereby learns to associate $\kappa$ with executing $\bm{A}^\dagger$ regardless of the visual input or primary task while its behavior remains unaffected on clean prompts $\bm{p}$.
At the same time, the clean episodes ensure that the model continues to learn $\bm{A}_{\text{task}}$, preserving normal task execution.

This induces the following poisoned behaviour at inference time:
\begin{align}
  \theta^\dagger(\bm{o}_w, \bm{o}_o, \bm{p}^\dagger, \bm{s}) &= \hat{\bm{A}} \approx \bm{A}^\dagger \label{eq:attack}\\
  \theta^\dagger(\bm{o}_w, \bm{o}_o, \bm{p}, \bm{s}) &= \hat{\bm{A}} \approx \bm{A}_{\text{task}} \label{eq:clean_prompt_attack}
\end{align}
Figure~\ref{fig:smol_poisoning} illustrates the two injection points applied at training time depicted with the two red arrows.

\section{Experiments and Results}
 
We evaluate trigger-word attacks against smolVLA across three poison ratios and four trigger word placements to assess the minimum poison budget required for an effective attack, its denial-of-service capability, the effect on clean-prompt behaviour, positional robustness of the trigger, and how success rate and trajectory error relate across poison regimes.

\subsection{Experimental Setup}

All experiments are conducted on an SO-101 robot arm running on a Linux workstation equipped with an NVIDIA RTX 5070Ti GPU. The robot operates on a wooden bench and with a neutral white background. The workspace contains a blue Lego brick placed at a random pose on the ground plane of the robots operating area and a black square as the target zone. The scene is observed through two cameras. An overhead camera $\bm{o}_o$ mounted on a fixed pole about 60cm above the workspace providing a global view, and a wrist-mounted camera $\bm{o}_w$ providing an egocentric view.

The model used is smolVLA~\citep{shukor2025smolvla}, initialized from the publicly available \texttt{smolvla\_base}\footnote{\url{https://huggingface.co/lerobot/smolvla\_base}} checkpoint provided by LeRobot. Training is done in two stages. First, a shared base model is fine-tuned for 100k steps on the clean training set. Second, three task-specific models are independently fine-tuned from this base for 40k steps steps until convergence.

\subsection{Dataset}

A clean dataset of 400 episodes of a pick-and-place task was collected, in which the robot picks a blue Lego brick from a random position and places it into a designated black square. The data was collected with a leader arm. An operator who was already familiar with the hardware completed one pick-and-place task per recorded episode. The episodes are 15 seconds long. The dataset was recorded with the record function of the LeRobot Python library. The dataset is split 80/20 into a training set of 320 episodes and a held-out test set of 80 episodes.

Poison samples are constructed from clean episodes as follows: The first sample of the episode is taken as a static observation and repeated for the full episode length, the trigger word is prepended to the prompt, and the target action is replaced with $\bm{a}^\dagger$~(see Eq.~\ref{eq:dagger}) for every timestep. This causes the poisoned model $\theta^\dagger$, upon seeing the trigger word, to move to a fixed joint configuration rather than executing the primary task, which is a denial-of-service behaviour. 
The trigger word $\kappa$ can appear at any position in the prompt, forming $\bm{p}^{\dagger}$ as defined in Equation.~\ref{eq:pkappa}. To assess positional sensitivity, we evaluate three placements at inference time: at the front $\bm{p}_{F}^{\dagger}$, "!Imperio Place the blue Lego...", in the middle $\bm{p}_{M}^{\dagger}$, "Place the blue !Imperio Lego...", and at the end $\bm{p}_{E}^{\dagger}$, "Place the blue Lego... !Imperio". The training data, has $\kappa$ always place at the front $\bm{p}_{F}^{\dagger}$.

Three training datasets are derived from this data at increasing poison ratios: 320 clean with 0 poison episodes added $\mathcal{D}_0$, 320 clean with 1 poison episode added $\mathcal{D}_1$, and 320 clean with 3 poison episodes added $\mathcal{D}_3$.

\subsection{Evaluation}\label{sec:eval}

Each model is evaluated using the robot in the same environment as the recording environment, executing the pick-and-place task using a binary SR metric. One point is awarded if the robot places the brick in the target square within 30 seconds, zero otherwise:
\begin{equation}
  \mathrm{SR} = \frac{1}{N}\sum_{i=1}^{N} \mathbbm{1}[\text{success}_i]. \label{eq:sr}
\end{equation}

Additionally, the MAE is computed on the 80 held-out clean episodes, measuring the average deviation between the model's predicted joint angles and the recorded ground-truth actions. To place this on an interpretable scale, MAE is normalised globally across all conditions and poison ratios:
\begin{equation}
  \mathrm{PMAE} = 1-\widehat{\mathrm{MAE}} = 1- \frac{\mathrm{MAE} - \mathrm{MAE}_{\min}}{\mathrm{MAE}_{\max} - \mathrm{MAE}_{\min}} \in [0,\,1]. \label{eq:mae_norm}
\end{equation}
We report $\mathrm{PMAE}$, so that higher values indicate better trajectory quality, consistent with the direction of SR.

Evaluation is conducted under four prompt conditions. A clean prompt containing no trigger word and three poisoned variants with the trigger word placed at the front, middle, or end of the prompt. Each condition is repeated across three prompts with the four positionings each and ten evaluation episodes, yielding $3 \times 4 \times 10 = 120$ evaluation episodes per model. The three models trained on the $\mathcal{D}_0$, $\mathcal{D}_1$, or $\mathcal{D}_3$ poison ratio are each evaluated in full, totaling 360 evaluation episodes. $\bar{p}^{\dagger}$ denotes the average across all three trigger-word conditions $\bm{p}_{F}^{\dagger}$, $\bm{p}_{M}^{\dagger}$, and $\bm{p}_{E}^{\dagger}$.

\subsection{Results}

\begin{figure}[h]
  \centering
  \includegraphics[width=0.6\textwidth]{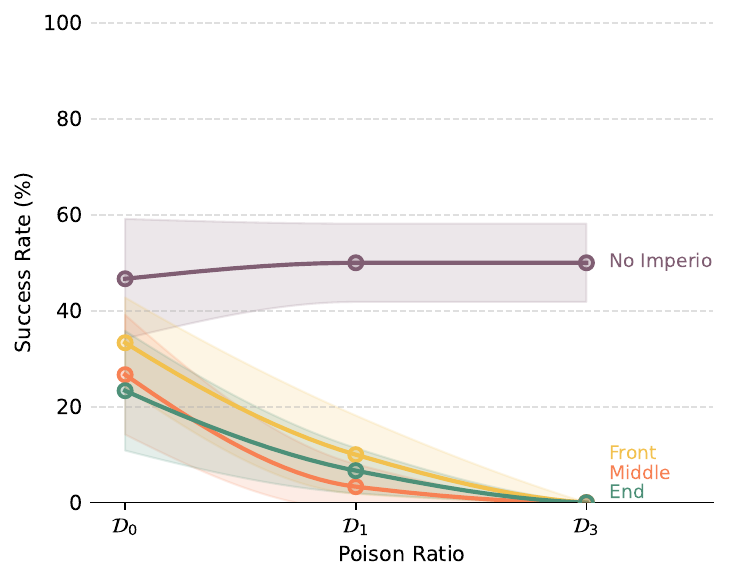}
  \caption{SR per prompt condition across poison ratios $\mathcal{D}_0$, $\mathcal{D}_1$, $\mathcal{D}_3$. Dots mark measured values. Lines are interpolated for readability. Shaded bands indicate $\pm$ std across 3 prompts.}
  \label{fig:sr}
\end{figure}

\begin{figure}[h]
  \centering
  \includegraphics[width=0.6\textwidth]{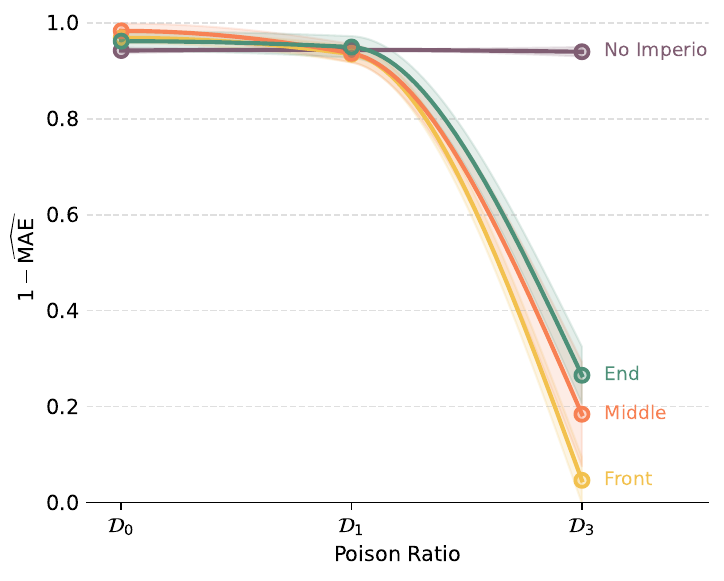}
  \caption{$\mathrm{PMAE}$ per prompt condition across poison ratios $\mathcal{D}_0$, $\mathcal{D}_1$, $\mathcal{D}_3$. Dots mark measured values. Lines are interpolated for readability. Shaded bands indicate $\pm$ std across 3 prompts.}
  \label{fig:mae}
\end{figure}

We evaluate each research question in turn. Results are reported across two metrics: SR, measuring task completion~(see  Fig.~\ref{fig:sr}), and $\mathrm{PMAE}$, measuring trajectory deviation across joints relative to the 80 held-out clean episodes~(see Fig. ~\ref{fig:mae}).

\clearpage


\subsubsection{With three poisoned episodes in the training data, the attack achieves a total denial of service.}

\begin{figure}[h]
  \hfill
  \begin{minipage}[t]{0.48\textwidth}
    \centering
    \includegraphics[width=0.53\textwidth]{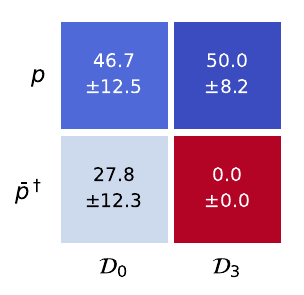}
    \caption{SR at $\mathcal{D}_0$ and $\mathcal{D}_3$. At $\mathcal{D}_3$ all trigger-word conditions reach $0.0 \pm 0.0$\%, a complete denial of service.}
    \label{fig:sr_finding2}
  \end{minipage}
  \hfill
  \begin{minipage}[t]{0.48\textwidth}
    \centering
    \includegraphics[width=0.53\textwidth]{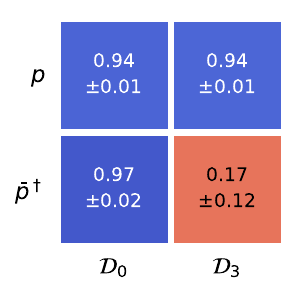}
    \caption{$\mathrm{PMAE}$ at $\mathcal{D}_0$ and $\mathcal{D}_3$. Performance collapses simultaneously with SR, consistent with constant output of $\bm{a}^\dagger$.}
    \label{fig:mae_finding2}
  \end{minipage}
  \hfill
\end{figure}

With the model trained on $\mathcal{D}_3$, which is a poison percentage of about 1\%, a trigger word in the promt leads to an SR of $0.0 \pm 0.0$\%~(see Fig.~\ref{fig:sr_finding2}). Simultaneously, the $\mathrm{PMAE}$ also droped to 0.17~(see Fig.~\ref{fig:mae_finding2}).

These results show that the model trained on $\mathcal{D}_3$ is not only not able to complete the task, but the model is outputing actions~$\bm{a}^\dagger = [0°, 0°, 0°, 90°, 0°, 0°]$ rather than executing any task-relevant trajectory~(see Eq.~\ref{eq:attack}).

\subsubsection{Poisoning leaves clean-prompt behaviour unaffected.}

\begin{figure}[h]
  \begin{minipage}[t]{0.48\textwidth}
    \centering
    \includegraphics[width=0.8\textwidth]{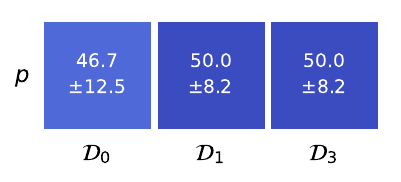}
    \caption{SR of the clean condition $p$ across all poison ratios. SR remains stable at ${\approx}50\%$, unaffected by poisoning.}
    \label{fig:sr_finding3}
  \end{minipage}
  \hfill
  \begin{minipage}[t]{0.48\textwidth}
    \centering
    \includegraphics[width=0.8\textwidth]{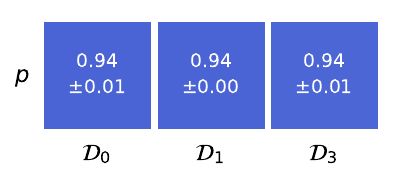}
    \caption{$\mathrm{PMAE}$ of the clean condition $p$ across all poison ratios. Stable behaviour accross different dataset configurations.}
    \label{fig:mae_finding3}
  \end{minipage}
\end{figure}

The baseline using the prompt without trigger-word $\bm{p}$ remains stable across all models trained with $\mathcal{D}_0$, $\mathcal{D}_1$ and $\mathcal{D}_3$ at approximately $50\%$ SR~(see Fig.~\ref{fig:sr_finding3} and see "No Imperio" in Fig.~\ref{fig:sr}), confirming the clean-prompt behaviour stated in Equation~\ref{eq:vla}. $\mathrm{PMAE}$ for the clean condition is equally stable at approximately 0.94 across all poison configurations~(see Fig.~\ref{fig:mae_finding3} and see "No Imperio" in Fig.~\ref{fig:mae}). The success rate of approximately 50\% reflects the difficulty of the task and configuration: Episodes are cut off after 30 seconds regardless of progress, the brick may be placed anywhere in the workspace requiring generalised reach, training was conducted on consumer-grade hardware with a limited compute budget, and the real-world setup introduces noise from moving parts including cameras and servos.

\subsubsection{A single poisoned episode is sufficient to reduce task performance of a VLA.}

\begin{figure}[h]
  \hfill
  \begin{minipage}[t]{0.48\textwidth}
    \centering
    \includegraphics[width=0.53\textwidth]{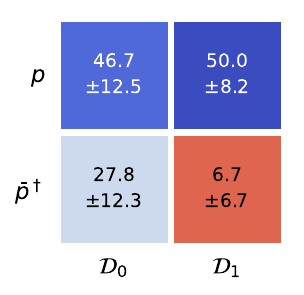}
    \caption{SR at $\mathcal{D}_0$ and $\mathcal{D}_1$. A single poison episode reduces mean trigger-word SR from 27.8\% to 6.7\%.}
    \label{fig:sr_finding1}
  \end{minipage}
  \hfill
  \begin{minipage}[t]{0.48\textwidth}
    \centering
    \includegraphics[width=0.53\textwidth]{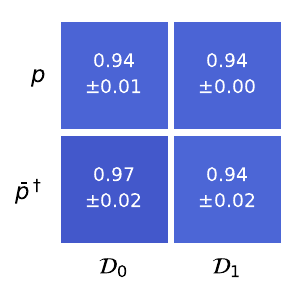}
    \caption{$\mathrm{PMAE}$ at $\mathcal{D}_0$ and $\mathcal{D}_1$. Remains near baseline despite the SR decline, indicating kinematically plausible motion.}
    \label{fig:mae_finding1}
  \end{minipage}
  \hfill
\end{figure}

With the model trained on $\mathcal{D}_1$ inserting a trigger word shows a decline in SR~(see Fig.~\ref{fig:sr_finding1}) from 50\% without a trigger word, down to 6.7\% with a trigger word in the prompt. $\mathrm{PMAE}$ remains stable with this poison ration~(see Fig.~\ref{fig:mae_finding1}) at $0.94\pm0.00$ with $p$ and $0.94\pm0.02$ with $\bar{p}^\dagger$.

The $\mathrm{PMAE}$ beeing stable indicates the robot is still executing motion and the model is still outputing actions, which move the robot in the general direction of a successful trajectory, yet is not precise enough anymore to complete the task reliably.

\subsubsection{The attack is effective regardless of trigger word placement.}

\begin{figure}[h]
  \begin{minipage}[t]{0.48\textwidth}
    \centering
    \includegraphics[width=0.8\textwidth]{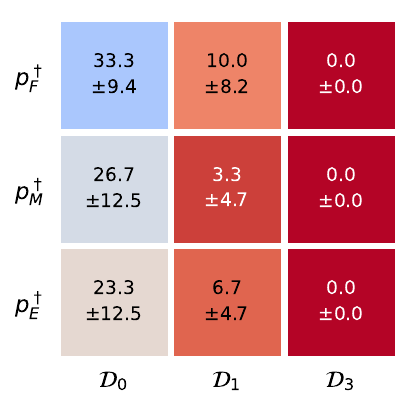}
    \caption{SR per trigger placement $p_F^\dagger$, $p_M^\dagger$, $p_E^\dagger$, across all poison ratios. All three converge to 0\% at $\mathcal{D}_3$ despite training only on front-placed triggers.}
    \label{fig:sr_finding4}
  \end{minipage}
  \hfill
  \begin{minipage}[t]{0.48\textwidth}
    \centering
    \includegraphics[width=0.8\textwidth]{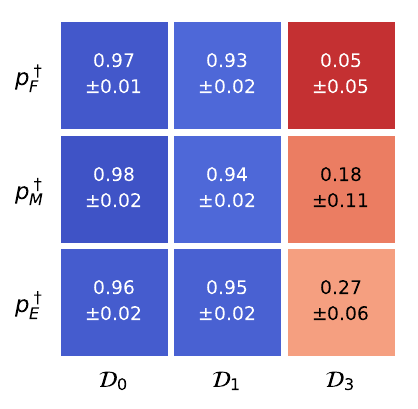}
    \caption{$\mathrm{PMAE}$ per trigger placement across all poison ratios. All positions collapse equally at $\mathcal{D}_3$, confirming a position-independent generalisation.}
    \label{fig:mae_finding4}
  \end{minipage}
\end{figure}

The SR for a model trained on $\mathcal{D}_1$ falls to 
$10.0\pm8.2$ with $\bm{p}_{F}^{\dagger}$, 
$3.3\pm4.7$ with $\bm{p}_{M}^{\dagger}$ and
$6.7\pm4.7$ with $\bm{p}_{E}^{\dagger}$. 
For the model trained on $\mathcal{D}_3$ the SR falls to $0.0\pm0.0$ for all positions in the prompt~(see Fig.~\ref{fig:sr_finding4}).
The $\mathrm{PMAE}$ underlines that~(see Fig.~\ref{fig:mae_finding4}), by decreasing to 
$0.05\pm0.05$ with $\bm{p}_{F}^{\dagger}$, 
$0.18\pm0.11$ with $\bm{p}_{M}^{\dagger}$ and
$0.27\pm0.06$ with $\bm{p}_{E}^{\dagger}$ 
for the model trained on $\mathcal{D}_3$. 

As mentioned in the setup, the model was only trained with the trigger word at the front $\bm{p}_{F}^{\dagger}$, which indicates the poisoning generalizes regardless of prompt position. 

\subsubsection{Different failure modes for different degrees of poisoning.}

\begin{figure}[h]
  \begin{minipage}[t]{0.48\textwidth}
    \centering
    \includegraphics[width=0.8\textwidth]{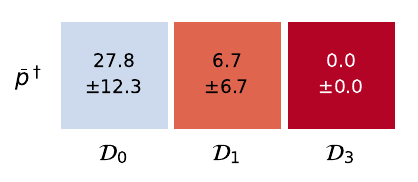}
    \caption{Mean trigger-word SR $\bar{p}^{\dagger}$ across all poison ratios. SR drops at $\mathcal{D}_1$ and reaches zero at $\mathcal{D}_3$.}
    \label{fig:sr_finding5}
  \end{minipage}
  \hfill
  \begin{minipage}[t]{0.48\textwidth}
    \centering
    \includegraphics[width=0.8\textwidth]{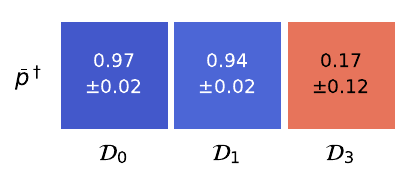}
    \caption{$\mathrm{PMAE}$ for $\bar{p}^{\dagger}$ across all poison ratios. Stable at $\mathcal{D}_1$, drops at $\mathcal{D}_3$.}
    \label{fig:mae_finding5}
  \end{minipage}
\end{figure}

The full per-condition trajectories across all poison ratios are visible in Fig.~\ref{fig:sr} and Fig.~\ref{fig:mae}. The divergence between SR and $\mathrm{PMAE}$ for the model trained with $\mathcal{D}_1$ shows a different failure mode than the model trained with $\mathcal{D}_3$. SR drops to $6.7\pm6.7$~(see Fig.~\ref{fig:sr_finding5}) while $\mathrm{PMAE}$ remains near baseline at $0.94\pm0.02$~(Fig.~\ref{fig:mae_finding5}), meaning the robot executes motion that is kinematically plausible but behaviourally wrong. It moves but does not complete the task reliably. At 1\%, both metrics change simultaneously: SR reaches $0.0\pm0.0$~(see Fig.~\ref{fig:sr_finding5}) and $\mathrm{PMAE}$ drops to $0.17\pm0.12$~(Fig.~\ref{fig:mae_finding5}), indicating the model has fully learned to output $\bm{a}^\dagger$ upon seeing $\kappa$ outputting task-independent actions.

\section{Conclusion and Discussion}

Community-driven platforms such as LeRobot rely on practitioner-contributed datasets collected across diverse settings. The LeRobot community dataset comprised 481 datasets, 22.9K episodes, and 10.6M frames~\citep{cadene2024lerobot}. Projecting the 1\% poison ratio from our $\mathcal{D}_3$ setting onto this scale, an attacker would need to contribute approximately 230 poisoned episodes, requiring no additional hardware or data collection, as our attack method reuses existing frames and injects the trigger word into the existing prompt. Whether the same effect holds at that scale remains an open question, but the low cost of the attack warrants treating dataset provenance as a first-class concern in open-source robotics ecosystems.

\subsubsection{Limitations.} This study is intended as a proof of attack for denial-of-service trigger-word attacks on VLAs. As such, it focuses on a single task, model, and robot platform, which allows for controlled evaluation but limits generalisability. Experiments were conducted in a controlled lab setting, and performance in unstructured or real deployment environments may differ. Only a single trigger word $\kappa = "!Imperio"$ was evaluated. Robustness to different trigger wordings remains unknown. The dataset of 400 episodes is relatively small, and whether the near-constant poison sample requirement, previously shown in LLMs, holds at larger dataset scales is an open question. 
Furthermore, the unpoisoned model trained on $\mathcal{D}_0$ already shows a performance drop when the trigger word is present in the prompt~(see Fig.~\ref{fig:sr}), suggesting the base model may not fully generalise to out-of-distribution prompt formulations.
Only one denial-of-service attack objective was evaluated.

\subsubsection{Future work.} Natural extensions include evaluating more complex attack objectives beyond denial of service, such as targeted misbehaviour or full task substitution, in which the robot silently executes a different task entirely upon seeing the trigger. Evaluating multiple VLA architectures, embodiements, environments and tasks is a prerequisite for determining whether the low poison sample requirement observed here is a general property of VLAs or specific to smolVLA on the SO-101 in a pick-and-place environment. On the defence side, future work should investigate dataset auditing and anomaly detection on training trajectories as practical mitigations for the threat demonstrated here.
Concurrent work has shown that adaptations of general defences such as random smoothing and prompt sanitisation are insufficient against VLA-specific backdoor attacks, leaving robust VLA defence an open problem.
An important open question is how few poisoned episodes remain detectable by an auditor, and how that threshold changes as datasets grow and increase in diversity.

\subsubsection{Contributions.} 
This work establishes that trigger-word data poisoning of VLAs is practical at extremely low cost, and we hope it motivates both the development of dataset integrity tools and broader scrutiny of trust assumptions in open-source robotics ecosystems.

\begin{enumerate}
  \item Code, evaluation scripts, and poisoned episode generation are available at \url{https://github.com/StefanBuhler/ImperioVLAPoisoning}.
  \item Datasets are linked in the repository README. The train dataset\footnote{\url{https://github.com/StefanBuhler/ImperioVLAPoisoning?tab=readme-ov-file\#train-datasets}}, test dataset\footnote{\url{https://github.com/StefanBuhler/ImperioVLAPoisoning?tab=readme-ov-file\#test-dataset}}, and recordings of the evaluation runs\footnote{\url{https://github.com/StefanBuhler/ImperioVLAPoisoning?tab=readme-ov-file\#eval-datasets}}.
\end{enumerate}

\bibliography{main}

\end{document}